\documentclass{article}

\usepackage[final]{neurips_2024}

\usepackage[utf8]{inputenc} 
\usepackage[T1]{fontenc}    
\usepackage{hyperref}       
\usepackage{url}            
\usepackage{booktabs}       
\usepackage{amsfonts}       
\usepackage{nicefrac}       
\usepackage{microtype}      
\usepackage{xcolor}         
\usepackage{subfigure}
\usepackage{graphicx}
\usepackage{amsmath}
\usepackage{amssymb}
\usepackage{enumitem}

\DeclareMathOperator*{\argmax}{argmax}
\definecolor{darkblue}{rgb}{0,0,0.88}

\title{Video Token Merging \\for Long-form Video Understanding}

\author{
  Seon-Ho Lee\thanks{Work done during an internship at Amazon Prime Video.} \\
  Korea University \\
  \texttt{seonholee@mcl.korea.ac.kr} \\
  \And
  Jue Wang\\
  Amazon AGI\\
  \texttt{juewangn@amazon.com}\\
  \AND
  Zhikang Zhang\\
  Amazon AGI\\
  \texttt{zhikang@amazon.com}\\
  \And
  David Fan\thanks{Work done while at Amazon Prime Video.}\\
  Meta FAIR\\
  \texttt{davidfan@meta.com} \\
  \And
  Xinyu Li \\
  Amazon AGI \\
  \texttt{xxnl@amazon.com} \\
}

\begin{document}

\maketitle

\begin{abstract}
  
As the scale of data and models for video understanding rapidly expand, handling long-form video input in transformer-based models presents a practical challenge. Rather than resorting to input sampling or token dropping, which may result in information loss, token merging shows promising results when used in collaboration with transformers. However, the application of token merging for long-form video processing is not trivial. We begin with the premise that token merging should not rely solely on the similarity of video tokens; the saliency of tokens should also be considered. To address this, we explore various video token merging strategies for long-form video classification, starting with a simple extension of image token merging, moving to region-concentrated merging, and finally proposing a learnable video token merging (VTM) algorithm that dynamically merges tokens based on their saliency. Extensive experimental results show that we achieve better or comparable performances on the LVU, COIN, and Breakfast datasets. Moreover, our approach significantly reduces memory costs by \textbf{84\%} and boosts throughput by approximately \textbf{6.89} times compared to baseline algorithms.
\end{abstract}

\section{Introduction}
\label{sec:intro}

Over the past few years, the Transformer architecture~\citep{vaswani2017attention} has risen as a revolutionary paradigm within natural language processing (NLP)~\citep{devlin2018bert} and has seamlessly expanded its influence into the domain of computer vision~\citep{wang2022deformable, bertasius2021space, wang2022long, akbari2021vatt, li2022mvitv2,fan2021multiscale}. This expansion has been exemplified by remarkable achievements in recent multi-modality foundation models such as Sora~\citep{videoworldsimulators2024}, GPT4~\citep{achiam2023gpt}, and Gemini~\citep{gemini}, showcasing its exceptional performance and versatility across diverse applications. 

\begin{figure}[t]
    \centering
    \includegraphics[width=0.75\linewidth]{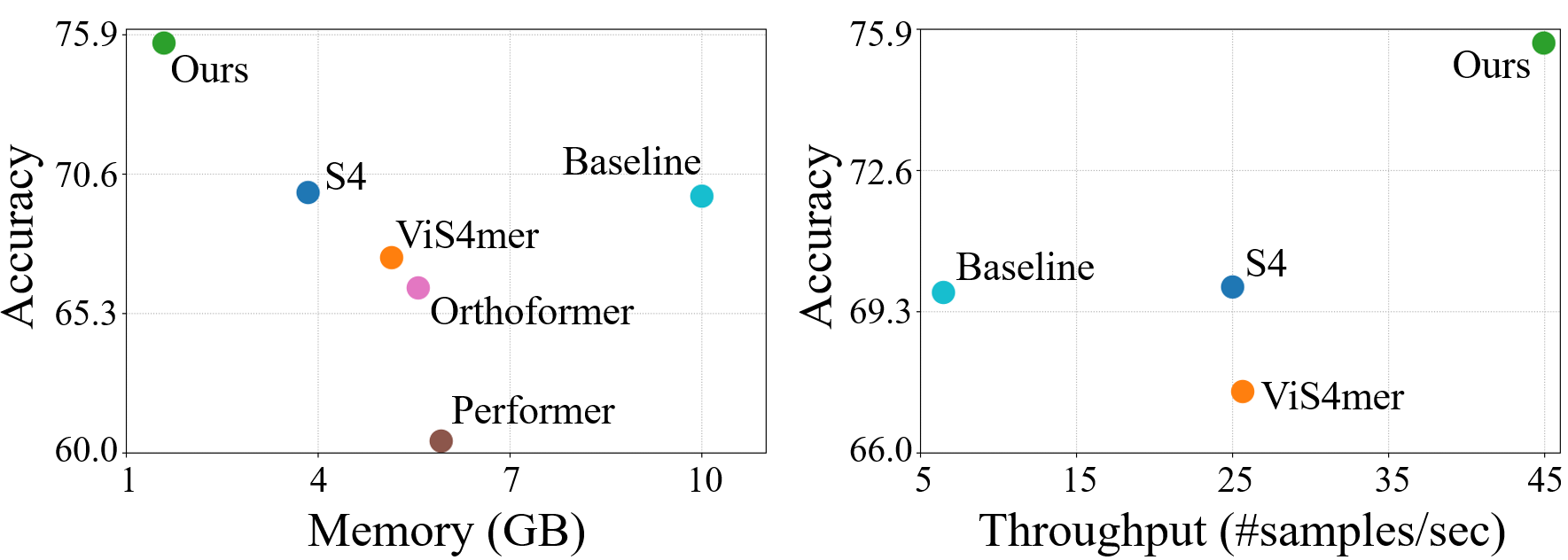}
    \caption{Comparison of GPU memory footprint and throughput against scene prediction accuracy on the LVU dataset~\citep{wu2021lvu}.}
    \label{fig:intro}
\end{figure}

In contrast to the natural language processing, the visual input has much lower information density and thus tokenizing the raw RGB image as non-overlapped patches becomes the essential operation in vision transformers~\citep{wang2022deformable, bertasius2021space, dosovitskiy2020vit}. However, the computational cost of transformer exponentially increases with the sequence length, which generates tremendous computation when feeding visual input into the large-scaled transformer models with billions of parameters. Due to the redundancy in video sequence, this phenomenon becomes more severe with video input, especially for long-form videos. This bottleneck impedes the further advancement of foundational models in handling long-form video data. Various attempts~\citep{yin2021adavit,wang2021efficient,meng2022adavit,rao2021dynamicvit,liang2022not} have been proposed to improve the efficiency of vision transformer by introducing a token selection module. However, these methods are primarily designed for images and may require non-trivial adaptation to the long-form video scenarios due to the video-level long-term dependencies and motion dynamics. Moreover, tokens dropped by the token selection module cannot be reused in later layers, which may result in the loss of important information.

In addition to the token selection, token merging techniques~\citep{bolya2022tome, ren2023testa, bolya2023tomesd, li2023vidtome} have been proposed to increase the efficiency and effectiveness of transformer-based networks. Specifically, they reduce the sequence length by merging similar tokens, thereby decreasing the computational cost. In addition to the efficiency, token merging demonstrates huge advantages by increasing the contextual information so that the model can learn from patterns presented across multiple tokens. Previous token merging algorithms in both the image and video domains use manually designed token partitioning methods and merge tokens based on their similarity. Even though the merged tokens would still keep the original information, they may have different granularity after the merging operation. In this paper, it is argued that different regions in the visual data may have different information density. Since the discriminative information of tokens may be degenerated after merging, some visual tokens should not be merged even if they look similar to each other. Rather than relying solely on similarity, we question whether more unmerged tokens should be used to describe salient areas, while merging more tokens for the background.

In this paper, we explore various video token merging (VTM) methods in long-form video classification task and aim to find the effective token merging method for long-form videos. Previous video token merging method~\citep{li2023vidtome} only decouples the spatial and temporal dimensions, which is unfavorable, especially for long-form videos. As the long-term dependency plays an important role in the long-form video understanding, spatiotemporal visual tokens should be considered jointly. Sequentially merging token along with one dimension after another may generate biased prior. In our work, we first naively extend the image-based token merging~\citep{bolya2022tome} to the video domain and then propose region-centralized and motion-based token merging algorithms, which estimate the salient region of video sequences. Finally, we develop a learnable VTM which predicts the saliency score of each token and adaptively merges spatiotemporal visual tokens in data-driven manner. Experimental results demonstrate that the proposed algorithm improves the effectiveness and the efficiency of the transformer-based network and outperforms the conventional long-video understanding methods with better throughput and less memory usage, as also shown in Figure~\ref{fig:intro}.

We summarize the contributions of this paper as following:
\begin{itemize}[align=parleft,left=10pt..1em, leftmargin=*]
\itemsep0em
    \item We explore various video token merging methods including the na\"ive VTM, the region-concentrated VTM, and the motion-based VTM.
    \item We propose the learnable video token merging algorithm, which estimates the saliency score of each token and adaptively merges visual tokens based on their scores. 
    \item The proposed algorithm achieves the best or competitive results on various datasets including LVU, Breakfast and COIN. Moreover, we significantly reduce memory costs by 84\% and improve the throughput by 6.89 times via the proposed learnable VTM.
\end{itemize}

\vspace*{-0.2cm}
\section{Related Work}
\label{sec:rel}
\vspace*{-0.2cm}
\subsection{Long-form Video Modeling} 
\vspace*{-0.2cm}
Transformers have demonstrated remarkable prowess in capturing long-term dependencies, as evidenced in their success in natural language processing (NLP) tasks~\citep{brown2020language,dai2019transformer,devlin2018bert}. However, the intensive computational requirements stemming from dense self-attention calculations~\citep{vaswani2017attention} pose a significant obstacle not only in NLP but also in the domain of computer vision, especially for the long-form videos. Many recent video transformer works~\citep{wang2022deformable, liu2021swin,bertasius2021space} focuses on improving the global attention mechanism. However, they are not designed for dealing with redundant spatial and temporal image tokens that are common in long-form video scenarios. To capture longer temporal information, LF-VILA~\citep{sunlong2022} develops a hierarchical architecture to include more frames in the model. Similarly, MeMViT~\citep{wu2022memvit} utilizes longer temporal information by emerging the previously cached ``memory" from the past. A novel alternative to transformers is the Structured State-Space Sequence (S4) model proposed by~\citet{gu2021efficiently}, which models the long-range dependencies by simulating a linear time invariant (LTI) system. Subsequently, ViS4mer~\citep{islam2022vis4mer} and S5~\citep{wang2023s5} extend S4 model to the long-form video classification task. ViS4mer~\citep{islam2022vis4mer} stacks multiple S4 layers with different scales in modeling long-form videos, and S5~\citep{wang2023s5} include an additional selective module to further improve the performance. Unlike these works that focus on the improvement of architecture and attention mechanism, this paper will start from a more basic concept in the transformer, video tokens and how to effectively merge them. Even though our proposed method can theoretically be applied on S4 model, the scope of this paper is on the well established transformer architecture. We will leave the investigation of video token merging on S4~\citep{gu2021efficiently} in the future work.

\vspace*{-0.2cm}
\subsection{Adaptive Token Selection} 
\vspace*{-0.2cm}
Adaptive token selection is widely used to improve model efficiency by leveraging a light-weight selection module to pick up the `useful' tokens while dropping the `unuseful' ones. In vision transformer, STTS~\citep{wang2021efficient} utilizes a token selection module known as the named scorer network to assign importance scores to each token, subsequently selecting the top-K frames with the highest scores. Building upon this concept, AdaViT ~\citep{meng2022adavit} further extends the approach by developing instance-specific policies. These policies guide the activation of patches, self-attention heads, and transformer blocks, enhancing adaptability and efficiency in processing visual data. STTS, AdaVit and other similar approaches~\citep{wang2021efficient,meng2022adavit,rao2021dynamicvit,liang2022not} drop a significant number of tokens in the early decision stage to save more cost, but the dropped tokens cannot be reused in the later layers, which is easier to degenerate the contextual information in the long-form videos. 

\vspace*{-0.2cm}
\subsection{Token Merging} 
\vspace*{-0.2cm}
Visual token merging is first proposed in~\citep{bolya2022tome} which aims at increasing the throughput of existing ViT models without training. Following works~\citep{ren2023testa, bolya2023tomesd, li2023vidtome} leverage this idea to save computational cost in different downstream applications, such as diffusion model, video and language understanding, and video editing. Specifically, visual tokens are first partitioned into two sets with equal size; for each of the edge tokens in one set, find the most similar token in another set and merge them by average pooling; finally, concatenate two sets back together. Although the token merging is simple and effective, its applications have mostly remained in the image domain. There is no fundamental research work has been explored for the long-form video token merging strategies, where the spatiotemporal tokens are more redundant and embed complicated dependencies locally and globally. In this work, we argue that visual tokens from the long-form video should be carefully partitioned and merged based on the salient areas in videos. To this end, we ablate various video token merging algorithms and provide extensive expermental results and analysis.

\vspace*{-0.3cm}
\section{Proposed Algorithm}
\label{sec:proposed}

\vspace*{-0.3cm}
\subsection{Preliminary -- Token Merging}
\vspace*{-0.2cm}
Token merging~\citep{bolya2022tome} aims to reduce the redundancy by merging similar tokens at each transformer block, thereby increasing the effectiveness and efficiency of a transformer-based network. Specifically, token merging has three steps: partitioning, matching, and merging. 

\vspace{0.1cm}
\noindent\textbf{Partitioning:} For given a set of $N$ tokens ${\cal X} = \{x_1, x_2, \ldots, x_N\}$, token merging first partition ${\cal X}$ into a set of target tokens $\cal T$ and a set of source tokens $\cal S$, given by 
\begin{eqnarray} 
{\cal T} &=& \{x_i : i \bmod \gamma = 0\}, \label{eq:partition_t} \\ 
{\cal S} &=& \{x_j : j \bmod \gamma \neq 0\} \label{eq:partition_s}
\end{eqnarray}
where $\gamma$ is partition factor and $\bmod$ denotes the modulo operator. Thus, $|{\cal T}| = \frac{|{\cal X}|}{\gamma}$. Also, ${\cal X} = {\cal T} \cup {\cal S}$ and ${\cal T} \cap {\cal S} = \varnothing$. 

\vspace{0.1cm}
\noindent\textbf{Matching:} Then, for each source token in $\cal S$, it finds the most similar target token in $\cal T$. Here, the similarity between tokens $x_i$ and $x_j$ is defined as the cosine similarity of the corresponding key vectors $k_i$ and $k_j$, which are obtained in the most recent self-attention layer. For a source token $x_j \in {\cal S}$, the index of its matched target token $m_j$ is obtained by,
\begin{equation} \label{eq:matching} 
m_j = \argmax_{i: \{i : x_i \in {\cal T}\}} {\frac{k^t_ik_j}{|k_i||k_j|}}.
\end{equation}

\vspace{0.1cm}
\noindent\textbf{Merging:} Lastly, token merging merges the tokens based on the matching results. For each target token $x_i \in {\cal T}$, it obtains the merged token $y_i$ by using average pooling,
\begin{equation} \label{eq:merging} 
    y_i = \frac{x_i + \sum_{j \in {\cal M}_i}{x_j}}{1 + |{\cal M}_i|} 
\end{equation}
where ${\cal M}_i = \{j : m_j = i, \forall x_j \in {\cal S}\}$ is the index set of source tokens which are matched with $x_i$. In this case, the number of tokens is reduced by $|\cal S|$ after token merging. It can also control the number of reduced tokens by $R$ by reassigning the matching index as
\begin{equation}
    m_j = -1
\end{equation}
for all $x_j$ except for the source tokens with the $R$ highest similarity scores. 

\subsection{Problem Definition}
\vspace*{-0.2cm}
Suppose that a video with $L$ frames is given, where $L \geq 60$. To perform the classification or regression on the given video, we can use a simple transformer-based network, which is shown in Figure~\ref{fig:architectures} (a) and (b). It first obtains the tokens $X_1, X_2, \ldots, X_L \in \mathbb{R}^{H \times W \times C}$ by using an encoder. Here, $H, W,$ and $C$ denote the height, the width, and the channel dimension of the token tensor, respectively. Then, it utilizes transformer blocks to update the tokens. As its input, $i$-th transformer block takes the tokens corresponding to $L_i$ frames without overlapping, where $L_i \leq L_j \leq L$ for $i < j$. The prediction head yields the estimation results. However, it requires the prohibitively large memory and computation costs due to the quadratic complexity of the self-attention ${\cal O}(L^2H^2W^2D^2)$, where $D$ is the dimension of key vectors. Hence, our goal is to increase the efficiency of this baseline network by reducing the number of tokens via token merging, while maintaining or even improving the performances of the network by removing the redundant or noisy information in the video. To this end, we explore various token merging methods for long video processing. 

\vspace*{-0.2cm}
\subsection{Video Token Merging -- Exploration}
\begin{figure}[t]
    \centering
    \includegraphics[width=0.85\linewidth]{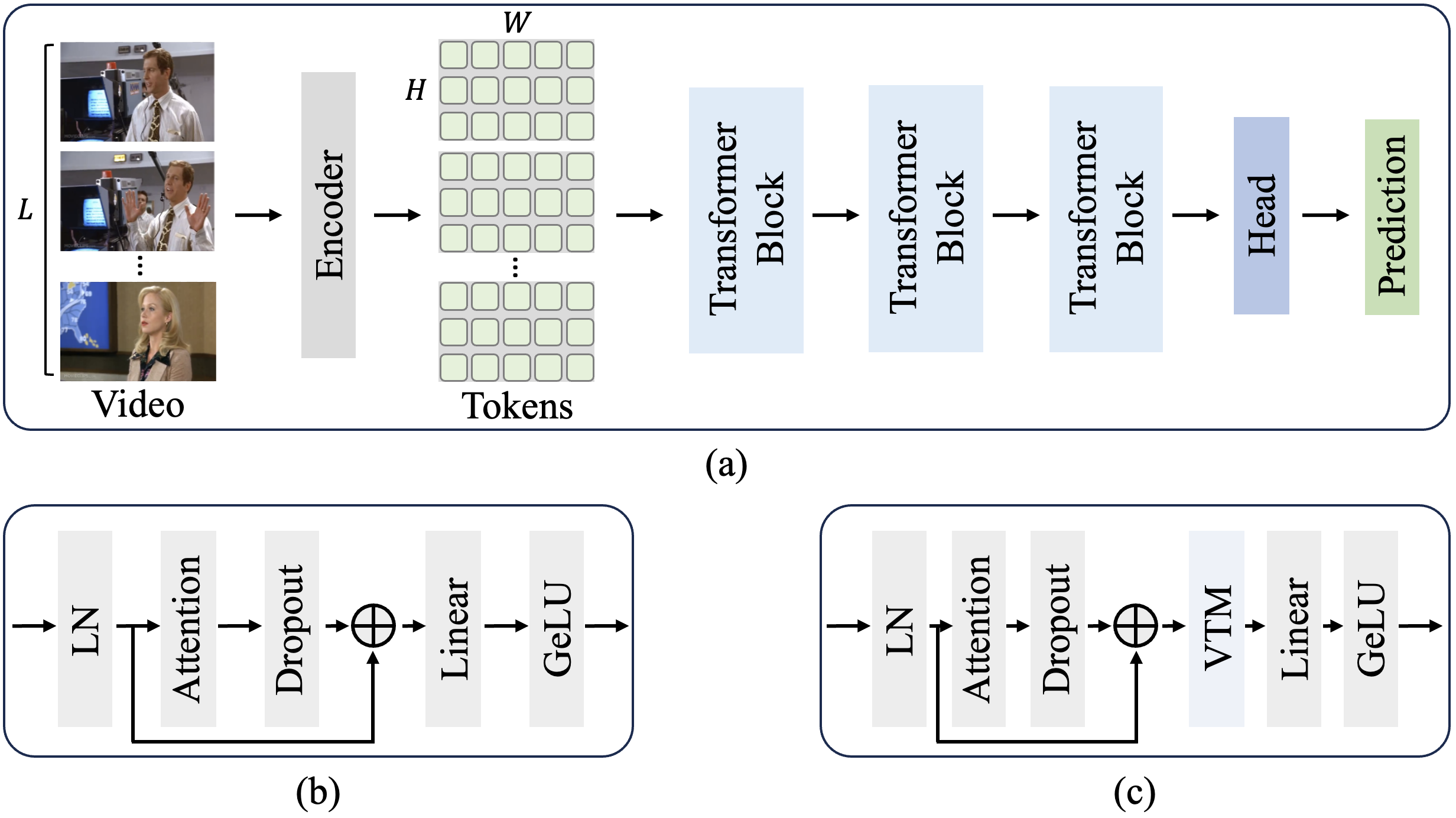}
    \caption{The architectures of (a) the baseline network, (b) the transformer block, and (c) the video token merging block.}
    \label{fig:architectures}
    \vspace*{-0.2cm}
\end{figure}

\vspace*{-0.2cm}
\noindent\textbf{Na\"ive video token merging:} First, we combine the standard token merging with the baseline network as intact as possible. To this end, we substitute transformer blocks with VTM blocks, in which token merging layer is inserted after the dropout layer, as illustrated in Figure~\ref{fig:architectures} (c). In this na\"ive VTM, we employ the standard token merging in~\eqref{eq:partition_t}-\eqref{eq:merging}. At each $i$-th VTM block, na\"ive VTM reduces the $R_i$ tokens. For example, at $\gamma = 4$ and $R =|\cal S|$, it gradually removes the $68\%$ of tokens over the network. Hence, the computation cost of the self-attention is reduced from ${\cal O}(N^2D^2)$ to ${\cal O}((\frac{N}{\gamma^{i-1}})^2D^2)$ at $i$-th VTM block. As shown in Table~\ref{table:vtm_comparison}, this na\"ive VTM shows better scores than the baseline network, because the token merging reduces the information redundancy in the videos.

\begin{figure}[t]
    \centering
    \includegraphics[width=\linewidth]{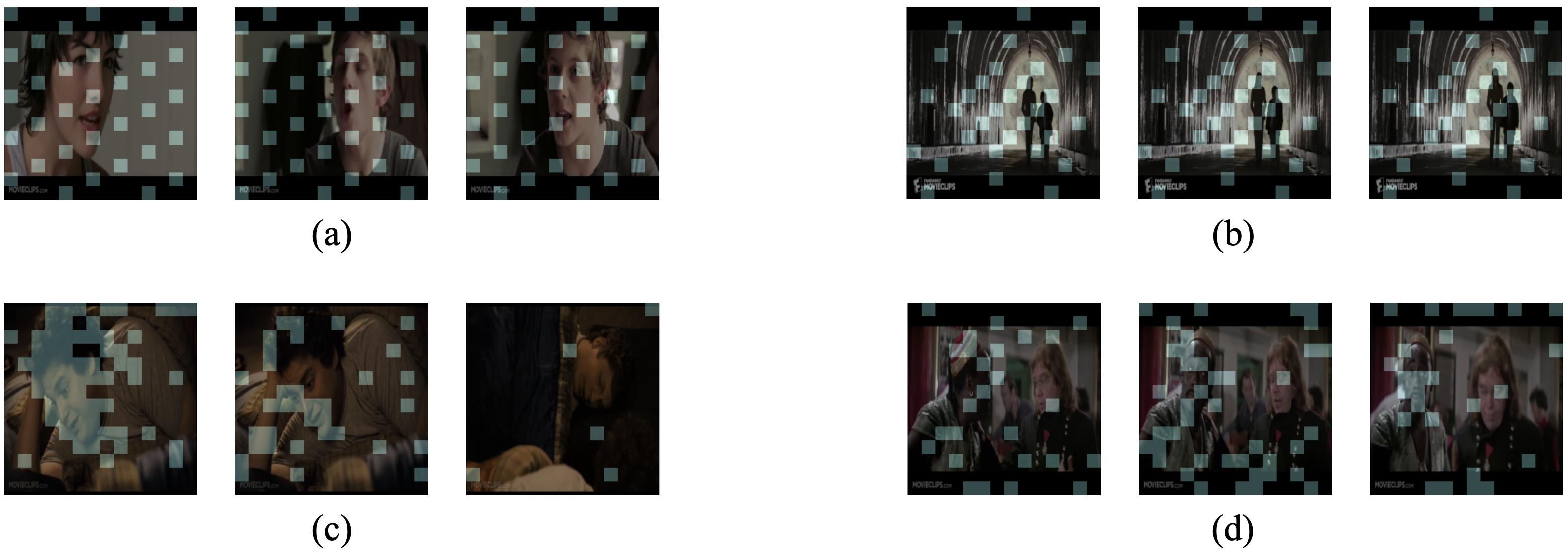}
    \vspace{-0.3cm}
    \caption{Visualizations of target tokens of different VTM methods: (a) na\"ive VTM, (b) center-concentrated VTM, (c) motion-based VTM, and (d) learnable VTM. In (d), learnable VTM selects the target tokens around salient objects rather than backgrounds.}
    \label{fig:target_examples}
    \vspace*{-0.3cm}
\end{figure}

\vspace{0.2cm}
\noindent\textbf{Region-concentrated video token merging:} Compared to an image, a video contains redundant spatiotemporal tokens. Depending on the tasks, some tokens are more important than others. However, as shown in Figure~\ref{fig:target_examples} (a), na\"ive VTM selects every $\gamma$-th token as the target tokens since it exploits uniform token partitioning in~\eqref{eq:partition_t}. Also, for token merging, it purely relies on the similarity between tokens regardless of the semantics, and thus the self-attention can more easily swayed by unnecessary information. Therefore, for better video token merging, it is important to consider the saliency of each token before merging them.

To investigate this issue, we explore center-concentrated video token merging and boundary-concentrated video token merging. The center-concentrated token merging samples $50\%$ of the entire target tokens from the center area with the size of $\frac{H}{2} \times \frac{W}{2}$, which uses more unmerged tokens to describe center area and merge more token from the boundary. Specifically, we use the partition factor of $\frac{\gamma}{2}$ for the center area and $\frac{3}{2}\gamma$ for the remaining area. On the other side, we implement the opposite operation for the boundary-concentrated video token merging. As shown in Table~\ref{table:vtm_comparison}, center-concentrated VTM shows better performances than na\"ive and boundary-concentrated VTM in general. Since meaningful objects and motions are typically center-concentrated, this suggests more tokens from salient regions should be unmerged while more of the rest tokens should be merged. Figure~\ref{fig:target_examples} (b) shows the partitioning results of center-concentrated VTM.

\begin{table}[t]
    \caption
    {
        Comparison of different VTM methods on the LVU dataset. The best results are boldfaced and the second-best ones are underlined.
    }
    \centering
    \resizebox{0.98\linewidth}{!}{
    \footnotesize
    \begin{tabular}{@{} l c c c c c c c c c c c @{} }
        \toprule
        && \multicolumn{3}{c}{Content $(\uparrow)$} & \multicolumn{4}{c}{Meta data $(\uparrow)$} & \multicolumn{2}{c}{User $(\downarrow)$} \\
        \cmidrule(lr){2-5} \cmidrule(lr){6-9} \cmidrule(lr){10-11}
        Algorithm && Relationship & Speaking & Scene & Director & Genre & Writer & Year & Like & View \\
        \midrule
          Baseline && 57.14 & 36.68 & 69.76 & 62.61 & 56.73 & 49.40 & 39.86 & 0.28 & 4.18 \\
          Na\"ive && \underline{61.90} & 36.18 & 72.09 & \underline{67.28} & 55.12 & 51.19 & 44.75 & 0.28 & \textbf{4.01} \\
          Boundary && 59.52 & 37.18 & 69.76 & 61.68 & 57.21 & 50.0 & \underline{47.55}  & 0.26 & 4.16  \\
          Center && \underline{61.90} & \underline{40.20} & \underline{74.41} & 62.61 & \underline{58.81} & 51.19 & 44.05 & 0.25 & \underline{4.11}  \\  
          Motion && \textbf{64.28} & 37.68 & \underline{74.41} & 64.48 & 58.49 & \textbf{55.95} & \underline{47.55} & \underline{0.24} & 4.13 \\    
          Learnable && \textbf{64.28} & \textbf{42.12} & \textbf{75.58} & \textbf{70.09} & \textbf{59.77} & \underline{53.57} & \textbf{48.55} & \textbf{0.21} & \textbf{4.01} \\ 

        \bottomrule
    \end{tabular}
    }
    \vspace*{-0.2cm}
   \label{table:vtm_comparison}
\end{table}

\vspace{0.2cm}
\noindent\textbf{Motion-based video token merging:} Even though center-concentrated VTM has shown better performances than na\"ive VTM, the meaningful tokens are not always located in the center area. Moreover, 
the hand-crafted partitioning method forces the center-concentrated VTM to select the same number of target tokens from each frame, which is not flexible enough when applied at scale. Therefore, we explore motion-based video token merging which divides tokens into $\cal T$ and $\cal S$ based on the motion information since the moving objects carry important cues in general~\citep{fan2023motion}. 

Let us assume that we have the magnitude of motion vector $v_i$ for each token $x_i$. We compute the sampling probability by using softmax
\begin{equation} \label{eq:softmax}
p_i = \frac{e^{v_i}}{\sum_{j=1}^Ne^{v_j}}.
\end{equation}
Note that the sampling probability is proportional to the motion magnitude. Then, we construct $\cal T$ by sampling $\frac{N}{\gamma}$ tokens from $\cal X$ with the sampling probability $p_i$ for each token $x_i$. 

The goal of VTM is to increase the efficiency of transformer-based network for long video understanding. Therefore, the motion information should be obtained with negligible time and computation costs. Hence, instead of estimating the motion information with an additional module, we use the motion information which is already stored in the video files; most modern video codecs, such as MPEG-4~\citep{richardson2004h}, H.264~\citep{richardson2004h}, and HEVC~\citep{wien2015high}, exploit the motion information for efficient compression. The motion decoding takes only 0.3 milliseconds for each frame which is negligible. Figure~\ref{fig:target_examples} (c) shows the token partitioning examples of motion-based VTM.


\vspace{-0.2cm}
\subsection{Learnable Video Token Merging}
\begin{figure}[t]
    \centering
    \includegraphics[width=0.92\linewidth]{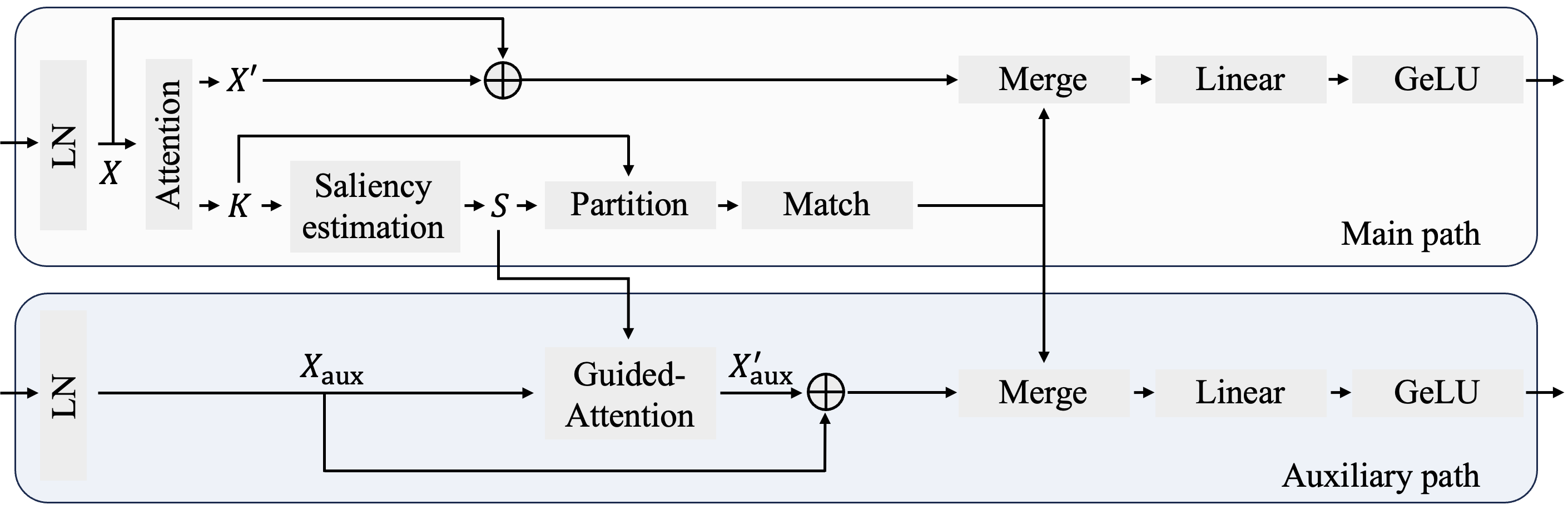}
    \vspace{-0.1cm}
    \caption{An overview of the learnable video token merging block. The auxiliary path is used during training only.}
    \label{fig:lrn_vtm}
    \vspace*{-0.1cm}
\end{figure}

\vspace{-0.2cm}
There are some videos in which unimportant objects have large motions due to various factors such as camera movement. Motion-based VTM may not yield good results on those videos. To maximize the generalizability, we develop learnable video token merging method. Instead of depending on the motion information, learnable VTM estimates the saliency score of each token and samples the target tokens based on the estimated scores. Figure~\ref{fig:lrn_vtm} shows the architecture of learnable VTM block. 

Learnable VTM block contains two forward paths: a main path and an auxiliary path. Let us assume that we have a tensor of $N$ tokens $X \in \mathbb{R}^{N \times C}$. 
In the main path, we first obtain query $Q$, key $K$, and value $V$ by
\begin{equation} \label{eq:qkv}
    Q = XU_q, \qquad K = XU_k, \qquad V = XU_v
\end{equation}
using learnable projection matrices $U_q, U_k, U_v \in \mathbb{R}^{C \times D}$. We perform the standard self-attention on $Q, K,$ and $V$ and yield the updated tokens $X'$ as
\begin{equation}
    X' = \frac{\mathrm{softmax}(QK^\top)}{\sqrt{D}}V .
\end{equation}
Also, from $K$, we estimate the saliency scores $S$ of tokens by  
\begin{equation}\label{eq:saliency}
S = [s_1, s_2, \ldots, s_N]^\top = \tanh({KU_s})
\end{equation}
where $U_s \in \mathbb{R}^{D \times 1}$ is a learnable matrix. Also, $s_i \in (-1, 1)$ for $1 \leq i \leq N$. Then, we compute the sampling probability using~\eqref{eq:softmax} with $s_i$ instead of $v_i$ for each token $x_i$ and sample $T$ with the sampling probability as in motion-based VTM. After the token partitioning, we match and merge the tokens in $X'$ by using~\eqref{eq:matching} and~\eqref{eq:merging}, respectively.

However, the learnable matrix $U_s$ in~\eqref{eq:saliency} can not be trained only with the main path since the partitioning process is not differentiable. To handle this issue, we employ the auxiliary path. This path consists of a saliency guided self-attention layer and a merging operation. The auxiliary path takes a tensor of auxiliary tokens $X_\mathrm{aux} \in \mathbb{R}^{N \times C}$ as its input. Similar to the main path, we obtain query $Q_\mathrm{aux}$, key $K_\mathrm{aux}$, and value $V_\mathrm{aux}$. Then, we perform the saliency guided attention to obtain the updated auxiliary tokens $X'_\mathrm{aux}$ as
\begin{equation}
    X'_\mathrm{aux} = \frac{\mathrm{softmax}(Q_\mathrm{aux}K_\mathrm{aux}^\top + \mathbf{1}S^\top)}{\sqrt{C}}V_\mathrm{aux} 
\end{equation}
where $\mathbf{1}$ is a $N$ dimensional vector of ones.
In the saliency guided attention, the contribution of each token is controlled by its estimated saliency score; if $s_i$ is positive, $i$-th token affects more on $X'_\mathrm{aux}$, whereas if $s_i$ is negative it has less influence on $X'_\mathrm{aux}$. In other words, it increases the influences of the tokens with high saliency scores in the attention process. Therefore, during training, the network is encouraged to assign high saliency scores to the tokens with meaningful information and low saliency scores to the others to obtain the better predictions. At the first VTM block, the auxiliary path employs the same input with the main path. From the second VTM block, it takes the output of auxiliary path in the previous VTM block as its input.

Also, it is worth pointing out that the auxiliary path is used for the network training only. Compared to other VTM methods, learnable VTM only introduces additional computation of score estimation module, which is fast and light enough, during test. Therefore, it shows almost same inference speed with other VTM methods. 

\vspace{-0.2cm}
\section{Experiments}
\vspace{-0.2cm}
\label{sec:exp}
\subsection{Datasets}

\noindent\textbf{LVU~\citep{wu2021lvu}:} It contains $\sim$30K videos sampled from $\sim$3K movies on the MovieClips~\citep{movieclip} website. Most videos are 1 to 3 minutes long. It provides the labels for 9 long-video understanding tasks which can be grouped into three major categories: 
\begin{itemize}[align=parleft,left=10pt..1em, leftmargin=*]\vspace{-0.15cm}
\itemsep0em
    \item Content understanding: `relationship,' `speaking style,' and `scene/place'
    \item Metadata prediction: `director,' `genre,' `writer,' and `movie release year'
    \item User engagement: `YouTube like ratio,' and `YouTube popularity'
\end{itemize} \vspace{-0.15cm}
As the evaluation metrics, we adopt the top 1 classification accuracy for content understanding and metadata prediction tasks and mean-squared error (MSE) for user engagement tasks.

\vspace{-0.1cm}
\noindent\textbf{Breakfast~\citep{kuehne2014breakfast}:} It provides 1,712 videos  with the average length of 2.32 minutes and the total length of 77 hours. The videos contain 52 individuals and 18 different backgrounds in total. Each video belongs to one of 10 complex cooking activities.

\vspace{-0.1cm}
\noindent\textbf{COIN~\citep{tang2019coin}:} It consists of 11,827 videos with the average length of 2.36 minutes, collected from YouTube. Each video belongs to one of 180 distinct procedural tasks. 

\begin{table}[t]
    \caption
    {
        Comparison of the proposed learnable VTM algorithm with conventional algorithms on the LVU dataset.
    }
    \centering
    \resizebox{0.99\linewidth}{!}{
    \footnotesize
    \begin{tabular}{@{} l c c c c c c c c c c c c c c @{} }
        \toprule
        && \multicolumn{3}{c}{Content $(\uparrow)$} & \multicolumn{4}{c}{Meta data $(\uparrow)$} & \multicolumn{2}{c}{User $(\downarrow)$} \\
        \cmidrule(lr){2-5} \cmidrule(lr){6-9} \cmidrule(lr){10-11}
        Algorithm && Relationship & Speaking & Scene & Director & Genre & Writer & Year & Like & View & GPU & Throughput\\
        \midrule
          Obj. T4mer~\citep{wu2021lvu} && 54.76 & 33.17 & 52.94 & 47.66 & 52.74 & 36.30 & 37.76 & 0.30 & 3.68  & - & - \\
          VideoBERT~\citep{sun2019videobert} && 52.80 & 37.90 & 54.90 & 47.30 & 51.90 & 38.50 & 36.10 & 0.32 & 4.46 & - & -\\
          Performer~\citep{choromanski2020performer} && 50.00 & 38.80 & 60.46 & 58.87 & 49.45 & 48.21 & 41.25 & 0.31 & 3.93 & 5.93GB & -\\
          Orthoformer~\citep{patrick2021orthoformer} && 50.00 & 38.30 & 66.27 & 55.14 & 55.79 & 47.02 & 43.35 & 0.29 & 3.86 & 5.56GB & -\\
          LST~\citep{islam2022vis4mer} && 52.38 & 37.31 & 62.79 & 56.07 & 52.70 & 42.26 & 39.16 & 0.31 & 3.83 & 41.38GB & -\\
          ViS4mer~\citep{islam2022vis4mer} && 57.14 & 40.79 & 67.44 & 62.61 & 54.71 & 48.80 & 44.75 & 0.26 & 3.63 & 5.15GB  &   25.64 \\
          S5~\citep{wang2023s5} && 61.98 & 41.75 & 69.88 & 66.40 & 58.80 & 50.60 & 47.70 & 0.25 & \textbf{3.51} & 3.85GB    &   25.0 \\
          S5+LSMCL~\citep{wang2023s5} && 61.98 & 41.75 & 72.53 & 66.40 & \textbf{61.34} & 50.60 & 47.70 & 0.24 & \textbf{3.51} & 3.85GB  &  25.0\\
          Learnable VTM && \textbf{64.28} & \textbf{42.12} & \textbf{75.58} & \textbf{70.09} & 59.77 & \textbf{53.57} & \textbf{48.55} & \textbf{0.21} & 4.01 & 1.60GB & 44.94\\

        \bottomrule
    \end{tabular}
    }
   \label{table:lvu}
\end{table}

\begin{table}[t]
    \caption
    {
       Comparison on the Breakfast dataset. PT stands for pretraining.
    }
    \centering
    \footnotesize
    \begin{tabular}{@{} l c c c c c @{} }
        \toprule
        Algorithm && PT Dataset & \#PT Samples & Accuracy \\
        \midrule
          VideoGraph~\citep{hussein2019videograph} && Kinetics-400 & 306K & 65.50  \\
          Timeception~\citep{hussein2019timeception} && Kinetics-400 & 136M & 71.30  \\
          GHRM~\citep{zhou2021graph} && Kinetics-400 & 495K &  75.50  \\
          D-sprv~\citep{lin2022learning} && HowTo100M & 136M &  89.90  \\
          ViS4mer~\citep{islam2022vis4mer} && Kinetics-600 & 495K & 88.17 \\
          S5~\citep{wang2023s5} && Kinetics-600 & 495K &  90.14   \\
          S5+LSMCL~\citep{wang2023s5} && Kinetics-600 & 495K &  90.70   \\          
          Learnable VTM && Kinetics-600 & 495K &  \textbf{91.26}  \\
        \bottomrule
    \end{tabular}
   \label{table:breakfast}
\end{table}

\subsection{Implementation Details}
\label{sec:detail}
We follow the experimental settings of conventional long-form video understanding algorithms~\citep{islam2022vis4mer, wang2023s5}. We employ three transformer blocks in the baseline network. As the encoder, we use ViT-L~\citep{dosovitskiy2020vit} pretrained on ImageNet-21K~\citep{ridnik2021imagenet} on the LVU~\citep{wu2021lvu} dataset and employ Swin-B~\citep{liu2021swin} pretrained on Kinetics-600~\citep{kay2017kinetics} on the Breakfast~\citep{kuehne2014breakfast} and COIN~\citep{tang2019coin} datasets. Images are resized to $224 \times 224$ for the feature extraction. Hence, $H=W=16$ and $C=1024$ for the LVU dataset and $H=W=7$ and $C=1024$ for the Breakfast and COIN datasets. The size of the length of input video for each dataset is also same with~\citep{islam2022vis4mer, wang2023s5}: we use 60 frames for the LVU dataset and 64 frames for the Breakfast and COIN datasets. We use
the AdamW~\citep{loshchilov2017decoupled} optimizer with a batch size of 16 and a weight decay of 0.01. We set the learning rate to 0.001. We train the network for 70 epochs by using cosine learning rate scheduler~\citep{gotmare2018closer} with 10 epochs warm-up. For experiments, we use 8 Tesla V100 GPUs and PyTorch.

\subsection{Experimental Results}
\label{sec:exp_results}
\noindent\textbf{Comparison on LVU:} In Table~\ref{table:lvu}, we compare the proposed algorithm with the conventional methods on the LVU dataset. With the smallest memory footprint, the proposed algorithm achieves the best scores in 7 out of 9 tasks on the LVU dataset. Performer~\citep{choromanski2020performer} and Orthoformer~\citep{patrick2021orthoformer} employ the efficient variants of self-attention to reduce the computation costs. The proposed learnable VTM outperforms these approaches with significant performance margins and less GPU memory usages. It shows the efficiency and effectiveness of our approach. Also, ViS4mer~\citep{islam2022vis4mer} and S5~\citep{wang2023s5} adopt S4 layers instead of self-attention layers to capture the long-term dependencies in videos because of its linear computation complexity to the number of input tokens. The promising results of these methods have suggested that S4 layer can be an efficient replacement of self-attention layer for long-form video inputs. However, the higher scores of the proposed algorithm broaden the potential usages of the self-attention layers for various long-form video tasks. Moreover, S5 utilizes LSMCL, which is a pretraining based on the contrastive learning, to boost its performances. Nevertheless, without the time-consuming pretraining, the proposed algorithm yields better scores on the LVU dataset.

\begin{table}[t]
    \caption
    {
       Comparison with the state-of-the-art methods on the COIN dataset. PT stands for pretraining. Here, $\ast$ means the reproduction results with the official codes.
    }
    \centering
    \footnotesize
    \begin{tabular}{@{} l c c c c c c @{} }
        \toprule
        Algorithm && PT Dataset & \#PT Samples & Accuracy \\
        \midrule
          TSN~\citep{tang2020comprehensive} && Kinetics-400 & 306K & 73.40  \\
          D-sprv~\citep{lin2022learning} && HowTo100M & 136M & 90.00  \\
          ViS4mer~\citep{islam2022vis4mer} && Kinetics-600 & 495K & 88.41 \\
          $\textrm{ViS4mer}^\ast$~\citep{islam2022vis4mer} && Kinetics-600 & 495K & 87.11 \\
          S5~\citep{wang2023s5} && Kinetics-600 & 495K & 90.42 \\
          S5+LSMCL~\citep{wang2023s5} && Kinetics-600 & 495K & \textbf{90.81} \\
          Learnable VTM && Kinetics-600 & 495K & 88.55 \\
        \bottomrule
    \end{tabular}
   \label{table:coin}
\end{table}

\begin{table}[t!]
\begin{minipage}[b]{0.48\linewidth}
\centering
\footnotesize
\caption
    {
        Comparison of long-video understanding results on the LVU dataset according to $\gamma$.
    }
\vspace*{-0.1cm}
\resizebox{\linewidth}{!}{
    \begin{tabular}{@{} l c c c c c c c c c c c @{} }
        \toprule
        Algorithm && Scene & Director & Like & Throughput & GPU\\
        \midrule
          Baseline  && 69.76 & 62.61 & 0.28 & 6.52 & 10GB \\
          $\gamma=2$  && 72.09 & 68.22 & 0.25 & 22.13 & 2.7GB \\
          $\gamma=6$  && 75.58 & 70.09 & 0.21 & 44.94 & 1.6GB \\
          $\gamma=10$ && 74.41 & 70.09 & 0.23 & 48.89 & 1.5GB \\ 
        \bottomrule
    \end{tabular}
    }
    \label{table:gamma_study}
\end{minipage}\hfill
\begin{minipage}[b]{0.49\linewidth}
\centering
\footnotesize
\caption
    {
       Comparison of long-video understanding results on the LVU dataset according to $(L_1, L_2, L_3)$.
    }
\vspace*{-0.18cm}
    \resizebox{\linewidth}{!}{
    \begin{tabular}{@{} l c c c c c c c c c c c @{} }
        \toprule
        $(L_1, L_2, L_3)$ && Scene & Director & Like & Throughput & GPU\\
        \midrule
        $(10, 30, 60)$ && 75.58 & 69.15 & 0.21 & 33.62 & 2.7GB \\
        $(6, 30, 60)$ && 75.58 & 70.09 & 0.21 & 44.94 & 1.6GB\\
        $(4, 20, 60)$ && 73.25 & 68.22 & 0.23 & 53.75 & 1.4GB \\
        \bottomrule
    \end{tabular} 
    }
    \label{table:L_study}
\end{minipage} 
\end{table}


\vspace{0.2cm}
\noindent\textbf{Comparison on Breakfast:} Table~\ref{table:breakfast} compares the performances of the proposed algorithm and the conventional techniques on the Breakfast dataset. The proposed learnable VTM achieves the best score on this challenging long-range activity classification dataset as well. For pretraining, D-sprv~\citep{lin2022learning} leverages the HowTo100M~\citep{miech2019howto100m} dataset which contains much more training samples than our pretraining dataset, Kinetics-600~\citep{carreira2018short}. Nevertheless, we outperform D-sprv~\citep{lin2022learning} with the accuracy gap of $1.36\%$.

\vspace{0.2cm}
\noindent\textbf{Comparison on COIN:} Table~\ref{table:coin} compares the scores of the proposed algorithm and the conventional techniques on the COIN dataset. We note that the COIN dataset consists of the videos on YouTube and more than 1,000 videos are not available anymore. Therefore, ViS4mer, which is one of the state-of-the-art method on the COIN dataset, achieves only $87.11\%$ accuracy when it is trained on the current version of the COIN dataset. It may be because of many missing training videos. Even though the comparison is not perfectly fair, we report the results on the COIN dataset for reference. The proposed learnable VTM yields better results than ViS4mer with the same training and test data. Also, it shows the comparable score with S5~\citep{wang2023s5}.







\begin{figure}[t]
    \centering
    \includegraphics[width=0.92\linewidth]{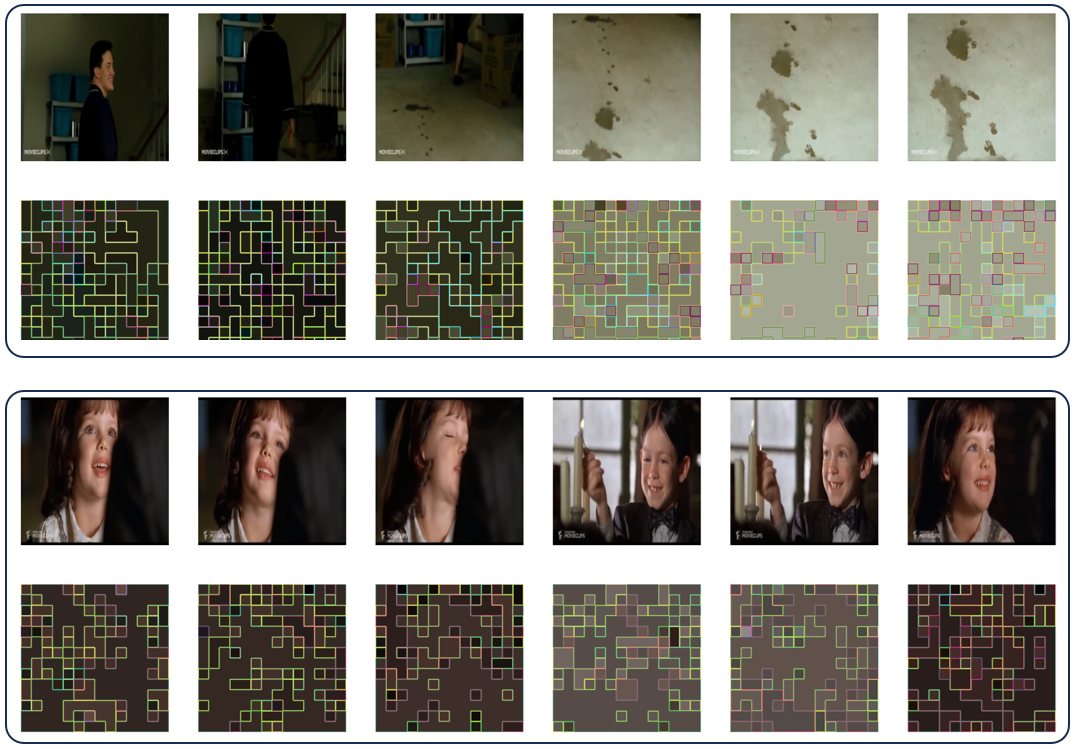}
    \vspace{-0.1cm}
    \caption{Visualizations of video token merging results on the LVU dataset. Patches with same inner and border color are merged together. The tokens corresponding to the backgrounds are merged together, thereby increasing the influence of salient tokens in the attention process.}
    \label{fig:exmaples}
    \vspace*{-0.1cm}
\end{figure}




\subsection{Analysis}

\noindent\textbf{Analysis on $\gamma$:} Table~\ref{table:gamma_study} compares the performances of learnable VTM with different $\gamma$. Compared to the baseline, at all $\gamma$, the proposed algorithm improves the performances. Also, at $\gamma=10$, it increase the throughput and the memory efficiency by 7.49 and 6.6 times, respectively. At $\gamma=6$, the proposed algorithm shows the best results. 

\vspace{0.2cm}
\noindent\textbf{Analysis on $(L_1, L_2, L_3)$:} Table~\ref{table:L_study} shows the results of the proposed learnable VTM with different $(L_1, L_2, L_3)$. Note that $i$-th VTM block takes the tokens in $L_i$ consecutive frames as its input at a time. At $(L_1, L_2, L_3) = (4, 20, 60)$, the proposed algorithm yields the lowest scores, since it can not capture the long temporal dependency in the early stage of the network. We see that the proposed algorithm yields the best scores at $(L_1, L_2, L_3) = (6, 30, 60)$. 

\vspace{0.2cm}
\noindent\textbf{Weighted average pooling:} To merge tokens, we use average pooling as the default setting in all VTM methods. However, once tokens are merged, they may represent more than one input patch. Thus, to reflect the token size in merging process, we combine tokens by averaging weighted by their sizes. However, as shown in Table~\ref{table:more_design_choices}, this weighted average pooling decrease the performances of learnable VTM. Thus, we exploit the average pooling to merge tokens.

\begin{table}[t]
\caption
    {
        Comparison of long-video understanding results of various VTM designs on the LVU dataset.
    }
\centering
\footnotesize

    \begin{tabular}{@{} l c c c c c c c c c c c @{} }
        \toprule
        Algorithm && Scene & Director & Like  \\
        \midrule
          Weighted average        && 72.09 & 68.22  & 0.24  \\
          Motion                  && 74.41 & 64.48  & 0.24  \\
          Motion weighted average && 74.41 & 66.35  & 0.23  \\
          Learnable               && 75.58 & 70.09  & 0.21 \\
          
        \bottomrule
    \end{tabular} 
    \label{table:more_design_choices}

\end{table}

\vspace{0.2cm}
\noindent\textbf{Motion weighted average pooling:} In motion-based VTM, we can combine tokens by averaging weighted by their motion magnitudes. Table~\ref{table:more_design_choices} shows the performances of motion-based VTM with the motion weighted average pooling. It yields the similar scores with the standard motion-based VTM with the average pooling. 

\vspace{0.2cm}
\noindent\textbf{Complexity:} Table~\ref{table:complexity} compares the throughput and memory footprint of learnable VTM during training and inference. Since the auxiliary path is additionally employed during training, it requires more computation costs. However, even during training, learnable VTM is still faster than conventional methods including ViS4mer~\citep{islam2022vis4mer} and S5~\citep{wang2023s5} and it also requires less amounts of memory than them.

\begin{table}[t]
\caption
    {
        Comparison of throughput and memory footprint of learnable VTM for training and inference.
    }
\centering
\footnotesize

    \begin{tabular}{@{} l c c c c c c c c c c c @{} }
        \toprule
        Algorithm && Phase & Throughput & Memory  \\
        \midrule
        ViS4mer~\citep{islam2022vis4mer} && Inference & 25.64 & 5.15GB \\
        S5~\citep{wang2023s5}      && Inference & 25 & 3.85GB \\
        Learnable VTM && Training & 27.84 & 2.8GB  \\ 
        Learnable VTM && Inference & 44.94 & 1.6GB \\
        \bottomrule
    \end{tabular} 
    \label{table:complexity}
    
\end{table}


\vspace{0.2cm}
\noindent\textbf{Visualizations:} In Figure~\ref{fig:exmaples}, we visualize the tokens merging results at the end of the network over multiple frames of video. We see that tokens with similar semantics are merged together. Also, tokens corresponding to backgrounds or unnecessary informations are merged more than tokens corresponding to salient objects. It is because learnable VTM selects tokens with high saliency scores as the target tokens. More visualization results are available in the supplemental document.

\vspace{-0.1cm}
\section{Conclusion}
\vspace{-0.1cm}
In this paper, we investigate the video token merging techniques for long-form video data. Unlike previous algorithms that apply uniform partitioning and merge tokens solely based on the visual similarity, we argue tokens with different saliencies should be treated unequally to avoid undesirable information loss after merging important tokens. To this end, we explore various video token merging methods and receive interesting intuitions from region-concentrated and motion-based token merging results. Lastly, we propose a learnable video token merging scheme that adaptively samples target tokens and learns discriminative representations from the long-form videos. Compared to the baseline, the proposed algorithm achieves substantial improvements in terms of the performance, memory cost and throughput.
\clearpage  

\bibliographystyle{iclr2024_conference}
\bibliography{main}
\clearpage

\appendix

\section{More Implementation Detail}
\label{sec:app_detail}
\subsection{Network Architecture}
\begin{figure*}[h!]
    \centering
    \includegraphics[width=\linewidth]{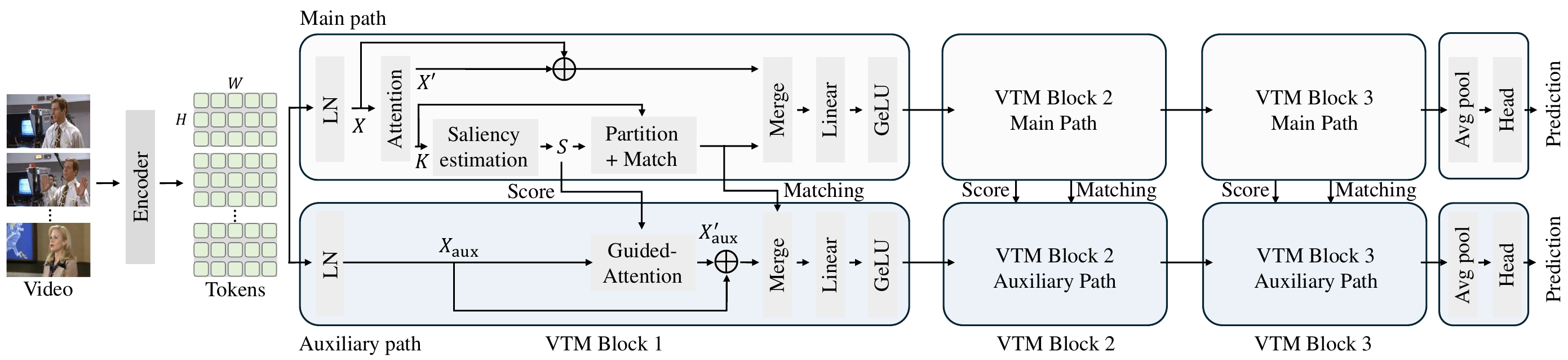}
    \caption{A network architecture of the proposed learnable VTM.}
    \vspace*{0.1cm}
    \label{fig:architecture}
    \vspace*{0.1cm}
\end{figure*}

Figure~\ref{fig:architecture} illustrates the detailed architecture of the proposed learnable VTM. For all datasets, the encoder extracts the tokens with 1024 channel dimension. We note that the linear layer in each VTM block reduces the channel dimension into half of it. Hence, after third VTM block, each token has 256 channel dimension. Also, the auxiliary path is only used for network training. 

\section{More Experiments}

\subsection{Analysis on $R$}
Table~\ref{table:R_study} compares the results of the proposed algorithm at different $R$. Note that $R$ denotes the number of merged tokens at each VTM block. At $R = |{\cal S}|$, we merge all source tokens with target tokens. However, there may exist some source tokens which does not have target tokens with similar semantics. Thus, at $R = |{\cal S}|$, undesirable merging of tokens can happen, thereby decreasing the performances. On the other hand, at $R = 0.5|{\cal S}|$, only the half of source tokens are merged with target tokens, and thus some source tokens may not be merged even though they have similar target tokens. Therefore, the proposed algorithm shows the best scores at $R = 0.8|{\cal S}|$.

\begin{table}[h!]
\caption
    {
        Comparison of long-video understanding results on the LVU dataset according to $R$.
    }
\centering
\footnotesize

    \begin{tabular}{@{} l c c c c c c c c c c c @{} }
        \toprule
        $R$ && Scene & Director & Like  \\
        \midrule
          $|{\cal S}|$    && 74.41 & 66.40 & 0.22  \\
          $0.8|{\cal S}|$ && 75.58 & 70.09 & 0.21  \\
          $0.5|{\cal S}|$ && 72.09 & 67.28 & 0.22  \\ 
        \bottomrule
    \end{tabular} 
    \label{table:R_study}

\end{table}

\clearpage
\section{More Visualizations}
Figure~\ref{fig:exmaples} visualizes the token merging results of the proposed learnable VTM on the LVU dataset.

\begin{figure}[h!]
    \centering
    \includegraphics[width=0.92\linewidth]{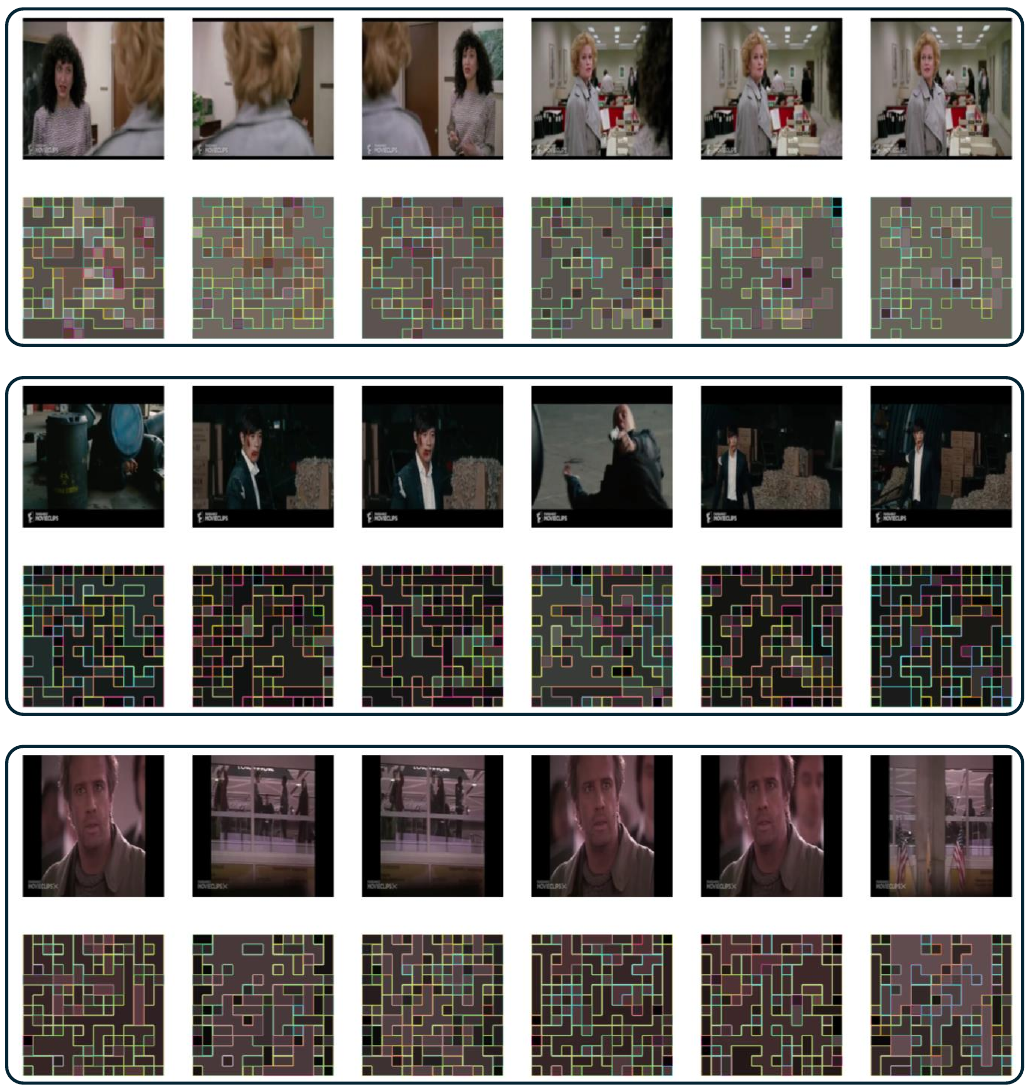}
    \caption{Visualizations of video token merging results on the LVU dataset. Patches with same inner and border color are merged together.}
    \label{fig:exmaples}
\end{figure}


\newpage
\section*{NeurIPS Paper Checklist}

\begin{enumerate}

\item {\bf Claims}
    \item[] Question: Do the main claims made in the abstract and introduction accurately reflect the paper's contributions and scope?
    \item[] Answer: \answerYes{} 
    \item[] Justification: See L65-L71 in Section~\ref{sec:intro}.
    \item[] Guidelines:
    \begin{itemize}
        \item The answer NA means that the abstract and introduction do not include the claims made in the paper.
        \item The abstract and/or introduction should clearly state the claims made, including the contributions made in the paper and important assumptions and limitations. A No or NA answer to this question will not be perceived well by the reviewers. 
        \item The claims made should match theoretical and experimental results, and reflect how much the results can be expected to generalize to other settings. 
        \item It is fine to include aspirational goals as motivation as long as it is clear that these goals are not attained by the paper. 
    \end{itemize}

\item {\bf Limitations}
    \item[] Question: Does the paper discuss the limitations of the work performed by the authors?
    \item[] Answer: \answerYes{} 
    \item[] Justification: See L280-287 in Section~\ref{sec:exp_results}.
    \item[] Guidelines:
    \begin{itemize}
        \item The answer NA means that the paper has no limitation while the answer No means that the paper has limitations, but those are not discussed in the paper. 
        \item The authors are encouraged to create a separate "Limitations" section in their paper.
        \item The paper should point out any strong assumptions and how robust the results are to violations of these assumptions (e.g., independence assumptions, noiseless settings, model well-specification, asymptotic approximations only holding locally). The authors should reflect on how these assumptions might be violated in practice and what the implications would be.
        \item The authors should reflect on the scope of the claims made, e.g., if the approach was only tested on a few datasets or with a few runs. In general, empirical results often depend on implicit assumptions, which should be articulated.
        \item The authors should reflect on the factors that influence the performance of the approach. For example, a facial recognition algorithm may perform poorly when image resolution is low or images are taken in low lighting. Or a speech-to-text system might not be used reliably to provide closed captions for online lectures because it fails to handle technical jargon.
        \item The authors should discuss the computational efficiency of the proposed algorithms and how they scale with dataset size.
        \item If applicable, the authors should discuss possible limitations of their approach to address problems of privacy and fairness.
        \item While the authors might fear that complete honesty about limitations might be used by reviewers as grounds for rejection, a worse outcome might be that reviewers discover limitations that aren't acknowledged in the paper. The authors should use their best judgment and recognize that individual actions in favor of transparency play an important role in developing norms that preserve the integrity of the community. Reviewers will be specifically instructed to not penalize honesty concerning limitations.
    \end{itemize}

\item {\bf Theory Assumptions and Proofs}
    \item[] Question: For each theoretical result, does the paper provide the full set of assumptions and a complete (and correct) proof?
    \item[] Answer: \answerNA{} 
    \item[] Justification: NA
    \item[] Guidelines:
    \begin{itemize}
        \item The answer NA means that the paper does not include theoretical results. 
        \item All the theorems, formulas, and proofs in the paper should be numbered and cross-referenced.
        \item All assumptions should be clearly stated or referenced in the statement of any theorems.
        \item The proofs can either appear in the main paper or the supplemental material, but if they appear in the supplemental material, the authors are encouraged to provide a short proof sketch to provide intuition. 
        \item Inversely, any informal proof provided in the core of the paper should be complemented by formal proofs provided in appendix or supplemental material.
        \item Theorems and Lemmas that the proof relies upon should be properly referenced. 
    \end{itemize}

    \item {\bf Experimental Result Reproducibility}
    \item[] Question: Does the paper fully disclose all the information needed to reproduce the main experimental results of the paper to the extent that it affects the main claims and/or conclusions of the paper (regardless of whether the code and data are provided or not)?
    \item[] Answer: \answerYes{} 
    \item[] Justification: See Section~\ref{sec:detail} and Appendix ~\ref{sec:app_detail}.
    \item[] Guidelines:
    \begin{itemize}
        \item The answer NA means that the paper does not include experiments.
        \item If the paper includes experiments, a No answer to this question will not be perceived well by the reviewers: Making the paper reproducible is important, regardless of whether the code and data are provided or not.
        \item If the contribution is a dataset and/or model, the authors should describe the steps taken to make their results reproducible or verifiable. 
        \item Depending on the contribution, reproducibility can be accomplished in various ways. For example, if the contribution is a novel architecture, describing the architecture fully might suffice, or if the contribution is a specific model and empirical evaluation, it may be necessary to either make it possible for others to replicate the model with the same dataset, or provide access to the model. In general. releasing code and data is often one good way to accomplish this, but reproducibility can also be provided via detailed instructions for how to replicate the results, access to a hosted model (e.g., in the case of a large language model), releasing of a model checkpoint, or other means that are appropriate to the research performed.
        \item While NeurIPS does not require releasing code, the conference does require all submissions to provide some reasonable avenue for reproducibility, which may depend on the nature of the contribution. For example
        \begin{enumerate}
            \item If the contribution is primarily a new algorithm, the paper should make it clear how to reproduce that algorithm.
            \item If the contribution is primarily a new model architecture, the paper should describe the architecture clearly and fully.
            \item If the contribution is a new model (e.g., a large language model), then there should either be a way to access this model for reproducing the results or a way to reproduce the model (e.g., with an open-source dataset or instructions for how to construct the dataset).
            \item We recognize that reproducibility may be tricky in some cases, in which case authors are welcome to describe the particular way they provide for reproducibility. In the case of closed-source models, it may be that access to the model is limited in some way (e.g., to registered users), but it should be possible for other researchers to have some path to reproducing or verifying the results.
        \end{enumerate}
    \end{itemize}

\item {\bf Open access to data and code}
    \item[] Question: Does the paper provide open access to the data and code, with sufficient instructions to faithfully reproduce the main experimental results, as described in supplemental material?
    \item[] Answer: \answerNo{} 
    \item[] Justification: NA
    \item[] Guidelines:
    \begin{itemize}
        \item The answer NA means that paper does not include experiments requiring code.
        \item Please see the NeurIPS code and data submission guidelines (\url{https://nips.cc/public/guides/CodeSubmissionPolicy}) for more details.
        \item While we encourage the release of code and data, we understand that this might not be possible, so “No” is an acceptable answer. Papers cannot be rejected simply for not including code, unless this is central to the contribution (e.g., for a new open-source benchmark).
        \item The instructions should contain the exact command and environment needed to run to reproduce the results. See the NeurIPS code and data submission guidelines (\url{https://nips.cc/public/guides/CodeSubmissionPolicy}) for more details.
        \item The authors should provide instructions on data access and preparation, including how to access the raw data, preprocessed data, intermediate data, and generated data, etc.
        \item The authors should provide scripts to reproduce all experimental results for the new proposed method and baselines. If only a subset of experiments are reproducible, they should state which ones are omitted from the script and why.
        \item At submission time, to preserve anonymity, the authors should release anonymized versions (if applicable).
        \item Providing as much information as possible in supplemental material (appended to the paper) is recommended, but including URLs to data and code is permitted.
    \end{itemize}

\item {\bf Experimental Setting/Details}
    \item[] Question: Does the paper specify all the training and test details (e.g., data splits, hyperparameters, how they were chosen, type of optimizer, etc.) necessary to understand the results?
    \item[] Answer: \answerYes{} 
    \item[] Justification: See Section~\ref{sec:exp}.
    \item[] Guidelines:
    \begin{itemize}
        \item The answer NA means that the paper does not include experiments.
        \item The experimental setting should be presented in the core of the paper to a level of detail that is necessary to appreciate the results and make sense of them.
        \item The full details can be provided either with the code, in appendix, or as supplemental material.
    \end{itemize}

\item {\bf Experiment Statistical Significance}
    \item[] Question: Does the paper report error bars suitably and correctly defined or other appropriate information about the statistical significance of the experiments?
    \item[] Answer: \answerNo{} 
    \item[] Justification: NA
    \item[] Guidelines:
    \begin{itemize}
        \item The answer NA means that the paper does not include experiments.
        \item The authors should answer "Yes" if the results are accompanied by error bars, confidence intervals, or statistical significance tests, at least for the experiments that support the main claims of the paper.
        \item The factors of variability that the error bars are capturing should be clearly stated (for example, train/test split, initialization, random drawing of some parameter, or overall run with given experimental conditions).
        \item The method for calculating the error bars should be explained (closed form formula, call to a library function, bootstrap, etc.)
        \item The assumptions made should be given (e.g., Normally distributed errors).
        \item It should be clear whether the error bar is the standard deviation or the standard error of the mean.
        \item It is OK to report 1-sigma error bars, but one should state it. The authors should preferably report a 2-sigma error bar than state that they have a 96\% CI, if the hypothesis of Normality of errors is not verified.
        \item For asymmetric distributions, the authors should be careful not to show in tables or figures symmetric error bars that would yield results that are out of range (e.g. negative error rates).
        \item If error bars are reported in tables or plots, The authors should explain in the text how they were calculated and reference the corresponding figures or tables in the text.
    \end{itemize}

\item {\bf Experiments Compute Resources}
    \item[] Question: For each experiment, does the paper provide sufficient information on the computer resources (type of compute workers, memory, time of execution) needed to reproduce the experiments?
    \item[] Answer: \answerYes{} 
    \item[] Justification: See Section~\ref{sec:detail} and Appendix~\ref{sec:app_complexity}.
    \item[] Guidelines:
    \begin{itemize}
        \item The answer NA means that the paper does not include experiments.
        \item The paper should indicate the type of compute workers CPU or GPU, internal cluster, or cloud provider, including relevant memory and storage.
        \item The paper should provide the amount of compute required for each of the individual experimental runs as well as estimate the total compute. 
        \item The paper should disclose whether the full research project required more compute than the experiments reported in the paper (e.g., preliminary or failed experiments that didn't make it into the paper). 
    \end{itemize}
    
\item {\bf Code Of Ethics}
    \item[] Question: Does the research conducted in the paper conform, in every respect, with the NeurIPS Code of Ethics \url{https://neurips.cc/public/EthicsGuidelines}?
    \item[] Answer: \answerYes{} 
    \item[] Justification: NA
    \item[] Guidelines:
    \begin{itemize}
        \item The answer NA means that the authors have not reviewed the NeurIPS Code of Ethics.
        \item If the authors answer No, they should explain the special circumstances that require a deviation from the Code of Ethics.
        \item The authors should make sure to preserve anonymity (e.g., if there is a special consideration due to laws or regulations in their jurisdiction).
    \end{itemize}

\item {\bf Broader Impacts}
    \item[] Question: Does the paper discuss both potential positive societal impacts and negative societal impacts of the work performed?
    \item[] Answer: \answerNA{} 
    \item[] Justification: NA
    \item[] Guidelines:
    \begin{itemize}
        \item The answer NA means that there is no societal impact of the work performed.
        \item If the authors answer NA or No, they should explain why their work has no societal impact or why the paper does not address societal impact.
        \item Examples of negative societal impacts include potential malicious or unintended uses (e.g., disinformation, generating fake profiles, surveillance), fairness considerations (e.g., deployment of technologies that could make decisions that unfairly impact specific groups), privacy considerations, and security considerations.
        \item The conference expects that many papers will be foundational research and not tied to particular applications, let alone deployments. However, if there is a direct path to any negative applications, the authors should point it out. For example, it is legitimate to point out that an improvement in the quality of generative models could be used to generate deepfakes for disinformation. On the other hand, it is not needed to point out that a generic algorithm for optimizing neural networks could enable people to train models that generate Deepfakes faster.
        \item The authors should consider possible harms that could arise when the technology is being used as intended and functioning correctly, harms that could arise when the technology is being used as intended but gives incorrect results, and harms following from (intentional or unintentional) misuse of the technology.
        \item If there are negative societal impacts, the authors could also discuss possible mitigation strategies (e.g., gated release of models, providing defenses in addition to attacks, mechanisms for monitoring misuse, mechanisms to monitor how a system learns from feedback over time, improving the efficiency and accessibility of ML).
    \end{itemize}
    
\item {\bf Safeguards}
    \item[] Question: Does the paper describe safeguards that have been put in place for responsible release of data or models that have a high risk for misuse (e.g., pretrained language models, image generators, or scraped datasets)?
    \item[] Answer: \answerNA{} 
    \item[] Justification: NA
    \item[] Guidelines:
    \begin{itemize}
        \item The answer NA means that the paper poses no such risks.
        \item Released models that have a high risk for misuse or dual-use should be released with necessary safeguards to allow for controlled use of the model, for example by requiring that users adhere to usage guidelines or restrictions to access the model or implementing safety filters. 
        \item Datasets that have been scraped from the Internet could pose safety risks. The authors should describe how they avoided releasing unsafe images.
        \item We recognize that providing effective safeguards is challenging, and many papers do not require this, but we encourage authors to take this into account and make a best faith effort.
    \end{itemize}

\item {\bf Licenses for existing assets}
    \item[] Question: Are the creators or original owners of assets (e.g., code, data, models), used in the paper, properly credited and are the license and terms of use explicitly mentioned and properly respected?
    \item[] Answer: \answerYes{} 
    \item[] Justification: NA
    \item[] Guidelines:
    \begin{itemize}
        \item The answer NA means that the paper does not use existing assets.
        \item The authors should citep the original paper that produced the code package or dataset.
        \item The authors should state which version of the asset is used and, if possible, include a URL.
        \item The name of the license (e.g., CC-BY 4.0) should be included for each asset.
        \item For scraped data from a particular source (e.g., website), the copyright and terms of service of that source should be provided.
        \item If assets are released, the license, copyright information, and terms of use in the package should be provided. For popular datasets, \url{paperswithcode.com/datasets} has curated licenses for some datasets. Their licensing guide can help determine the license of a dataset.
        \item For existing datasets that are re-packaged, both the original license and the license of the derived asset (if it has changed) should be provided.
        \item If this information is not available online, the authors are encouraged to reach out to the asset's creators.
    \end{itemize}

\item {\bf New Assets}
    \item[] Question: Are new assets introduced in the paper well documented and is the documentation provided alongside the assets?
    \item[] Answer: \answerNA{} 
    \item[] Justification: NA
    \item[] Guidelines:
    \begin{itemize}
        \item The answer NA means that the paper does not release new assets.
        \item Researchers should communicate the details of the dataset/code/model as part of their submissions via structured templates. This includes details about training, license, limitations, etc. 
        \item The paper should discuss whether and how consent was obtained from people whose asset is used.
        \item At submission time, remember to anonymize your assets (if applicable). You can either create an anonymized URL or include an anonymized zip file.
    \end{itemize}

\item {\bf Crowdsourcing and Research with Human Subjects}
    \item[] Question: For crowdsourcing experiments and research with human subjects, does the paper include the full text of instructions given to participants and screenshots, if applicable, as well as details about compensation (if any)? 
    \item[] Answer: \answerNA{} 
    \item[] Justification: NA
    \item[] Guidelines:
    \begin{itemize}
        \item The answer NA means that the paper does not involve crowdsourcing nor research with human subjects.
        \item Including this information in the supplemental material is fine, but if the main contribution of the paper involves human subjects, then as much detail as possible should be included in the main paper. 
        \item According to the NeurIPS Code of Ethics, workers involved in data collection, curation, or other labor should be paid at least the minimum wage in the country of the data collector. 
    \end{itemize}

\item {\bf Institutional Review Board (IRB) Approvals or Equivalent for Research with Human Subjects}
    \item[] Question: Does the paper describe potential risks incurred by study participants, whether such risks were disclosed to the subjects, and whether Institutional Review Board (IRB) approvals (or an equivalent approval/review based on the requirements of your country or institution) were obtained?
    \item[] Answer: \answerNA{} 
    \item[] Justification: NA
    \item[] Guidelines:
    \begin{itemize}
        \item The answer NA means that the paper does not involve crowdsourcing nor research with human subjects.
        \item Depending on the country in which research is conducted, IRB approval (or equivalent) may be required for any human subjects research. If you obtained IRB approval, you should clearly state this in the paper. 
        \item We recognize that the procedures for this may vary significantly between institutions and locations, and we expect authors to adhere to the NeurIPS Code of Ethics and the guidelines for their institution. 
        \item For initial submissions, do not include any information that would break anonymity (if applicable), such as the institution conducting the review.
    \end{itemize}

\end{enumerate}

\end{document}